\documentclass[conference]{IEEEtran}
\usepackage{booktabs}
\usepackage{gensymb} 
\IEEEoverridecommandlockouts
\usepackage{cite}
\usepackage{amsmath,amssymb,amsfonts}
\usepackage{algorithmic}
\usepackage{graphicx}
\usepackage{textcomp}
\usepackage{xcolor}
\usepackage{epstopdf} 
\usepackage{float}  
\usepackage{placeins} 
\usepackage{url}

\def\BibTeX{{\rm B\kern-.05em{\sc i\kern-.025em b}\kern-.08em
    T\kern-.1667em\lower.7ex\hbox{E}\kern-.125emX}}

\makeatletter
\newcommand{\linebreakand}{%
  \end{@IEEEauthorhalign}
  \hfill\mbox{}\par
  \mbox{}\hfill\begin{@IEEEauthorhalign}
}
\makeatother

\begin{document}

\title{A Benchmark Reference for ESP32-CAM Module}

\author{
\IEEEauthorblockN{1\textsuperscript{st} Nowroz, Sayed T.}
\IEEEauthorblockA{\textit{Computer Science \& Engineering} \\
\textit{Wright State University}\\
Dayton, USA \\
tnowroz@gmail.com}
\and
\IEEEauthorblockN{2\textsuperscript{nd} Saleh, Nermeen M.}
\IEEEauthorblockA{\textit{Computer Science \& Engineering} \\
\textit{Wright State University}\\
Dayton, USA \\
saleh.57@wright.edu}
\and
\IEEEauthorblockN{3\textsuperscript{rd} Shakur, Siam}
\IEEEauthorblockA{\textit{Electrical \& Electronic Engineering} \\
\textit{Ahsanullah Univ. of Science \& Technology}\\
Dhaka, Bangladesh \\
shakursiam@gmail.com}
\linebreakand
\IEEEauthorblockN{4\textsuperscript{th} Banerjee, Sean}
\IEEEauthorblockA{\textit{Computer Science \& Enggineering} \\
\textit{Wright State University}\\
Dayton, USA \\
sean.banerjee@wright.edu}
\and
\IEEEauthorblockN{5\textsuperscript{th}Amsaad, Fathi}
\IEEEauthorblockA{\textit{Computer Science \& Enggineering} \\
\textit{Wright State University}\\
Dayton, USA \\
fathi.amsaad@wright.edu}}

\maketitle

\begin{abstract}
The ESP32-CAM is one of the most widely adopted open-source modules for prototyping embedded vision applications. Since its release in 2019, it has gained popularity among both hobbyists and professional developers due to its affordability, versatility, and integrated wireless capabilities. Despite its widespread use, comprehensive documentation of the performance metrics remains limited. This study addresses this gap by collecting and analyzing over six hours of real-time video streaming logs across all supported resolutions of the OV2640 image sensor, tested under five distinct voltage conditions via an HTTP-based WiFi connection. A long standing bug in the official Arduino ESP32 driver, responsible for inaccurate frame rate logging, was fixed. The resulting analysis includes key performance metrics such as instantaneous and average frame rate, total streamed data, transmission count, and internal chip temperature. The influence of varying power levels was evaluated to assess the reliability of the module. 
\end{abstract}

\begin{IEEEkeywords}
Internet of Things (IoT), Embedded Systems, ESP32-CAM, Frame Rate
\end{IEEEkeywords}

\section{Introduction}
Unlike general-purpose computers, embedded systems are optimized for real-time performance with power, speed, and memory constraints. Over the past decade, the rise of Internet-connected devices has transformed the consumer market, driven by advances in low-power computing and wireless networking \cite{Background_arduino}. Portable IoT applications exist across diverse sectors such as industrial automation, robotics, smart agriculture, space and deep-sea exploration, and healthcare. These systems are significantly enhanced with the integration of cameras, enabling advanced imaging and analysis in scenarios where traditional infrastructure-based computer vision systems may not be feasible. Recently, the demand for cost-effective and vision-enabled embedded systems has increased. 

In response to growing demand for affordable embedded vision solutions, devices like the ESP32-CAM have become widely adopted. This open-source module combines an OV2640 image sensor with an ESP32S microcontroller, offering compactness, wireless connectivity, and sufficient processing power to support IoT-based vision applications \cite{AIThinkerESP32Cam}. Despite its popularity, there is lack of documentation on performance benchmarks and thermal characteristics of this hardware. This paper addresses that gap by presenting a comprehensive experimental study with real data, evaluating key performance metrics of the ESP32-CAM. The study incorporates findings enabled by a recent bug fix related to incorrect frames per second (FPS) calculation in the Arduino ESP32 Core repository, ensuring the most accurate and up-to-date assessment of the module’s capabilities. The contribution of this study is as follows:

\begin{enumerate}
\item \textbf{Bug Fix:} A long-standing bug in the official ESP32 Arduino Core repository was resolved, which had caused inaccurate FPS reporting since 2018 \cite{GitHubPR123}.

\item \textbf{Public Release of Benchmark Dataset:} The authors published dataset containing timestamped video streaming logs (including FPS, temperature, and payload metrics) collected at 15-minute intervals in all test cases, available at GitHub \cite{esp32cambenchmark}.

\item \textbf{Comprehensive Benchmarking:} An exhaustive evaluation of frame rates in all supported resolutions of the ESP32-CAM was performed, establishing a detailed performance baseline for future design.
\end{enumerate}


\section{Background}

The ESP32-CAM is a compact and low-cost development board introduced by AI-Thinker in 2019. It integrates the ESP32S microcontroller, built-in WiFi connectivity, a microSD card slot for local storage, an LED flash, and multiple GPIO and peripheral interfaces operating at 240 MHz. The board typically interfaces with the OV2640 image sensor, a 2-megapixel camera capable of streaming video and capturing still images in various resolutions. The ESP32S microcontroller on board configures the image sensor via the I2C protocol, while the DVP interface delivers the encoded image in an electrical signal. Supported encodings include RGB565, RGB555, and JPEG, with JPEG compression enabling higher-resolution image capture (up to 1600x1200) at lower frame rates. In contrast, RGB formats are limited to lower resolutions (320x240) due to RAM constraints \cite{RAM_Limitation}.

The Arduino framework is an open source platform designed to simplify embedded system programming through a rich ecosystem of hardware libraries and cross-platform compatibility. It uses a high-level language based on C/C++ and provides developers with tools to compile, and upload firmware to various microcontrollers \cite{arduinobackground}. For ESP32-CAM, it exposes high-level APIs from the Espressif IoT Development Framework (ESP-IDF), streamlining integration of WiFi, Bluetooth, and camera functionalities \cite{ide}. Its accessibility makes it well-suited for rapid prototyping, educational projects, and lightweight commercial deployments.

Node-RED is an open-source, flow-based development tool built on Node.js, widely adopted for designing event-driven IoT applications \cite{nodered_about}. Its visual programming interface allows for connecting devices, APIs, and services using drag-and-drop nodes, minimizing the need for traditional coding. In this work, Node-RED is used to manage UART-based real-time video stream logs, perform string processing, and export data to CSV files. Its lightweight design and flexibility make it well-suited for real-time data handling in embedded system workflows like those involving the ESP32-CAM.


\section{Related Work}
The ESP32-CAM is a low-cost microcontroller with an integrated camera that has widespread adoption in domains such as agriculture, surveillance, healthcare, home automation, and more. Its open-source ecosystem and built-in WiFi have enabled rapid prototyping and deployment of vision-based systems for real-time monitoring. This section summarizes recent applications of ESP32-CAM across several sectors.

ESP32-CAM has been extensively used in agriculture automation due to its compact form and imaging capabilities. In \cite{siam-1}, it was used to continuously monitor poultry health by integrating machine learning and Internet of Things (IoT), providing early diagnosis using image-based data. Similarly, \cite{siam-2} presented a solar powered ESP32-CAM system for remote plant growth monitoring. It has also been applied in automated weed detection and elimination systems using image-based ML interface\cite{siam-3}. ESP32 has also been deployed for farm security to detect animals and human activity via pose estimation algorithms to detect intrusions \cite{siam-4}.

In IoT architectures, ESP32-CAM often serves as an edge device, transmitting image data to local or cloud-based AI systems for analysis. In \cite{Alzimer}, a wearable ESP32-CAM system for Alzheimer patients is proposed, streaming video to a Raspberry Pi running MobileNet and DenseNet models for face detection and recognition. A similar approach is seen in \cite{9842706}, where the module was used to detect citrus diseases using MobileNet. IoT-based home security systems have made use of ESP32-CAM, using facial recognition for real-time surveillance \cite{siam-5, siam-8}. The live video feed was streamed, triggered by human presence detection, with an immediate notification to the owner. Other IoT-based home security work uses real-time facial recognition for automated door access \cite{siam-6}, recognizing visitors, and notifying the homeowner with photos of unfamiliar faces.

ESP32 has enough resources to host lightweight AI models, optimized to perform in real-time with low power and memory footprint. \cite{ESP32-NN-Benchmark} has deployed variations of fully connected neural networks in the Xtensa core. In \cite{smart-city-inspection}, a model is implemented to accurately detect flood, fire, traffic accident, and normal conditions on 96x96 images with 850 ms inference time. Applications such as robotic face tracking and aerial disaster detection in drones \cite{drone-esp32} further showcase the potential of ESP32-CAM in TinyML use cases. ESP32-CAM has also been adopted in commercial products. Prusa, a very well-known company in the 3D Printer market, uses ESP32-CAM for real-time print monitoring \cite{ESP32CamGithub}. The trend in ESP32-CAM and vision-based IoT applications has been increasing almost exponentially since 2020. Multi-node IoT vision network and security have been another new trend in the field \cite{ghajari2025networkanomalydetectioniot}.


\section{Methodology}
 \begin{figure}
    \centering
    \includegraphics[width=0.9\linewidth]{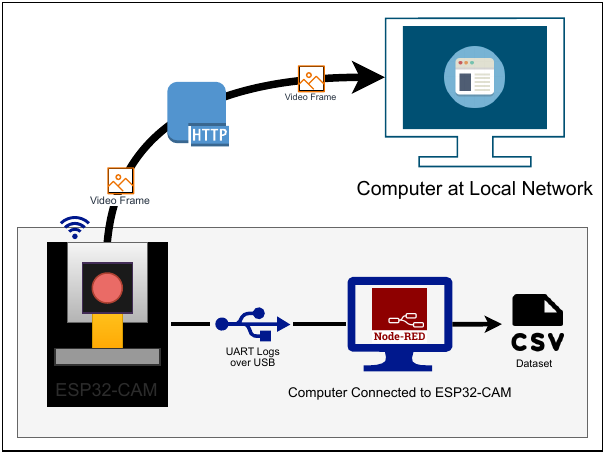}
    \caption{Experimental setup and implementation for real-time video streaming log collection.}
    \label{fig:enter-label}
\end{figure}

In this study, 15-minute video streaming logs were collected for each supported resolution of the ESP32-CAM module, with video streamed over WiFi via an HTTP server. A client can access the HTTP server hosted by ESP32-CAM on the local WiFi network, by accessing the IP address of the ESP32-CAM module. Real-time logs were transmitted from the ESP32-CAM through the UART serial interface and collected via USB. Node-RED was used to access, parse, and write logs into CSV files. For each transmitted video frame, a comma-separated log string was generated containing real-time statistics, including instantaneous FPS, average FPS, runtime, bytes transmitted, and chip temperature.

The firmware used is based upon Arduino's CameraServer example code, with corrected longstanding FPS logging bug in the example code. The firmware was modified to stream extra parameters such as temperature within the logs and dynamically control image resolution over UART. However, the resolution can also be changed from the HTTP interface. To study the impact of power variation, the same approach was adopted with controlled voltage levels applied directly to the ESP32-CAM power pin. UART logs were decoded and collected using a logic analyzer.

In total, approximately six hours of video streaming logs were collected, covering 15 available frame sizes and 8 distinct voltage levels. Each video frame transmission generated a timestamped entry containing the following features: runtime, bytes transmitted, instant frame time (ms), instant FPS, average frame time (ms), average FPS, temperature (°C), frame size, and voltage. The complete firmware, Node-RED flow, collected dataset, and data analysis tools are publicly available in the project's GitHub repository \cite{esp32cambenchmark}.


\section{Data Analysis and Observations}

\begin{table}[H]
    \centering
\caption{Average FPS and Temperature for Different Frame Sizes}
\label{tab:fps_temp}
    \begin{tabular}{|c|c|c|c|}
        \hline
        \textbf{Frame Size} & \textbf{Avg FPS} & \textbf{Avg Temperature (\degree C)} & \textbf{File Size (Bytes)} \\ \hline
        96x96 & 51.17 & 71.36 & 2003.09 \\ \hline 
        160x120 & 51.30 & 71.09 & 2924.12 \\ \hline 
        128x128 & 52.59 & 69.73 & 2005.48 \\ \hline 
        176x144 & 52.45 & 70.29 & 2881.65 \\ \hline 
        240x176 & 48.49 & 74.18 & 4929.59 \\ \hline 
        240x240 & 43.21 & 71.70 & 5107.69 \\ \hline 
        320x240 & 44.44 & 71.85 & 5947.81 \\ \hline 
        400x296 & 35.26 & 73.53 & 10614.46 \\ \hline 
        480x320 & 20.87 & 70.48 & 10015.03 \\ \hline 
        640x480 & 14.19 & 71.25 & 15572.34 \\ \hline 
        800x600 & 8.25 & 72.34 & 26978.11 \\ \hline 
        1024x768 & 3.37 & 72.68 & 40524.96 \\ \hline 
        1280x720 & 1.99 & 70.33 & 46796.23 \\ \hline 
        1280x1024 & 2.27 & 71.68 & 67638.45 \\ \hline 
        1600x1200 & 1.29 & 71.62 & 108466.02 \\ \hline
    \end{tabular}
\end{table}

The Table-\ref{tab:fps_temp} presents the average FPS and temperature while streaming video over HTTP on a local network. A significant performance drop is observed at and beyond 320x240 resolution. The highest tested resolution, 1600×1200, records an FPS of just 1.29. 

Frame rate reductions stem from increased frame acquisition times, which are influenced by sensor readout, data transmission, and firmware processing. Many of the mentioned factors increase non-linearly with the increase of resolution. Higher resolutions demand longer sensor readout durations due to increased pixel count and introduce greater computational overhead in the image signal processor. Notably, JPEG compression involves discrete cosine transform (DCT) operations with $O(n^2)$ complexity, which contribute to the non-linear decline in FPS at high resolutions.

\begin{figure}[htbp]
    \centering
    \begin{minipage}[b]{0.45\textwidth}
        \centering
        \includegraphics[width=\textwidth]{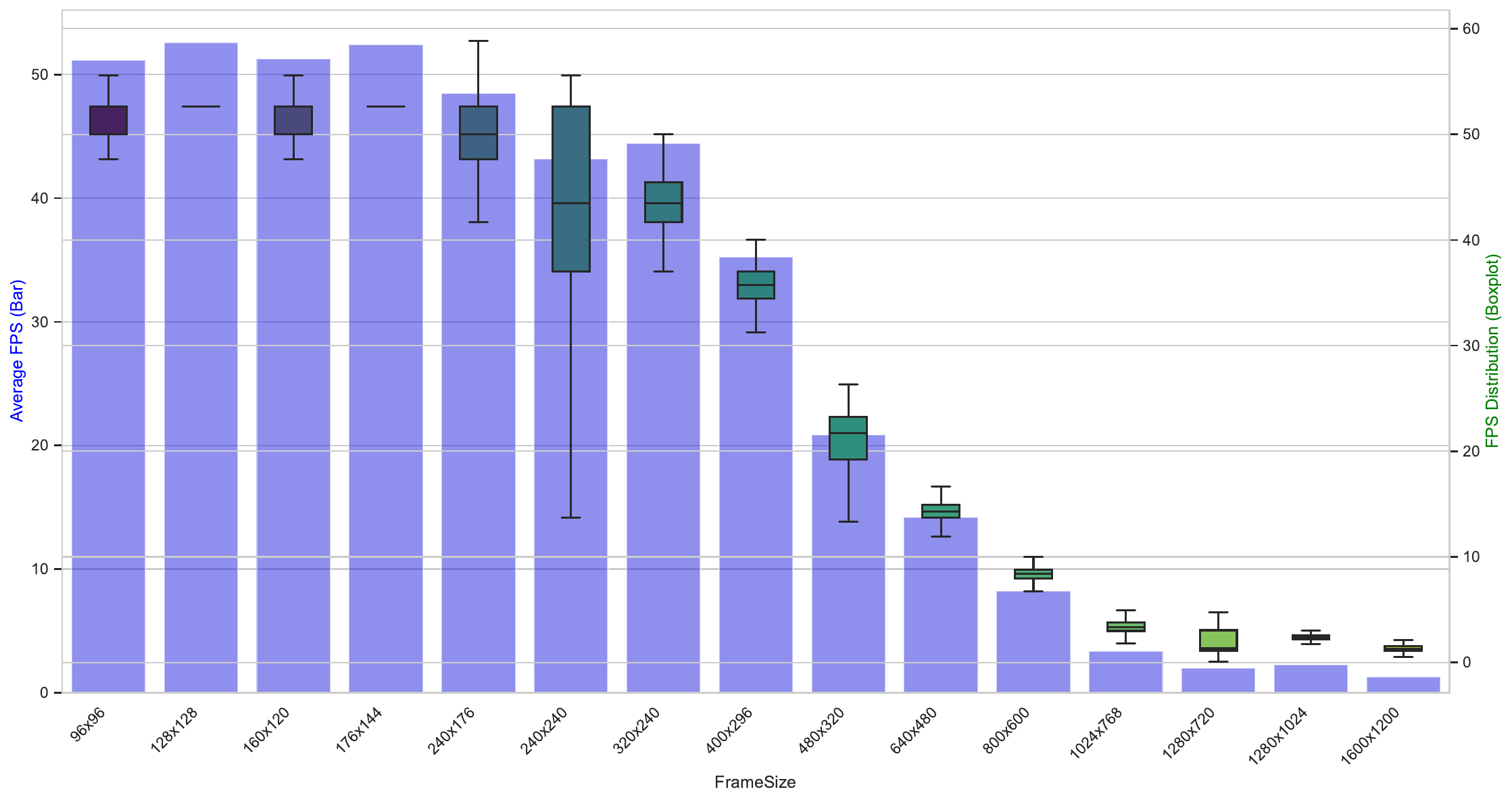}
        \caption{Average FPS and FPS Distribution for each Resolution.}
        \label{fig:frame_size_fps}
    \end{minipage}
    \hfill
    \begin{minipage}[b]{0.45\textwidth}
        \centering
        \includegraphics[width=\textwidth]{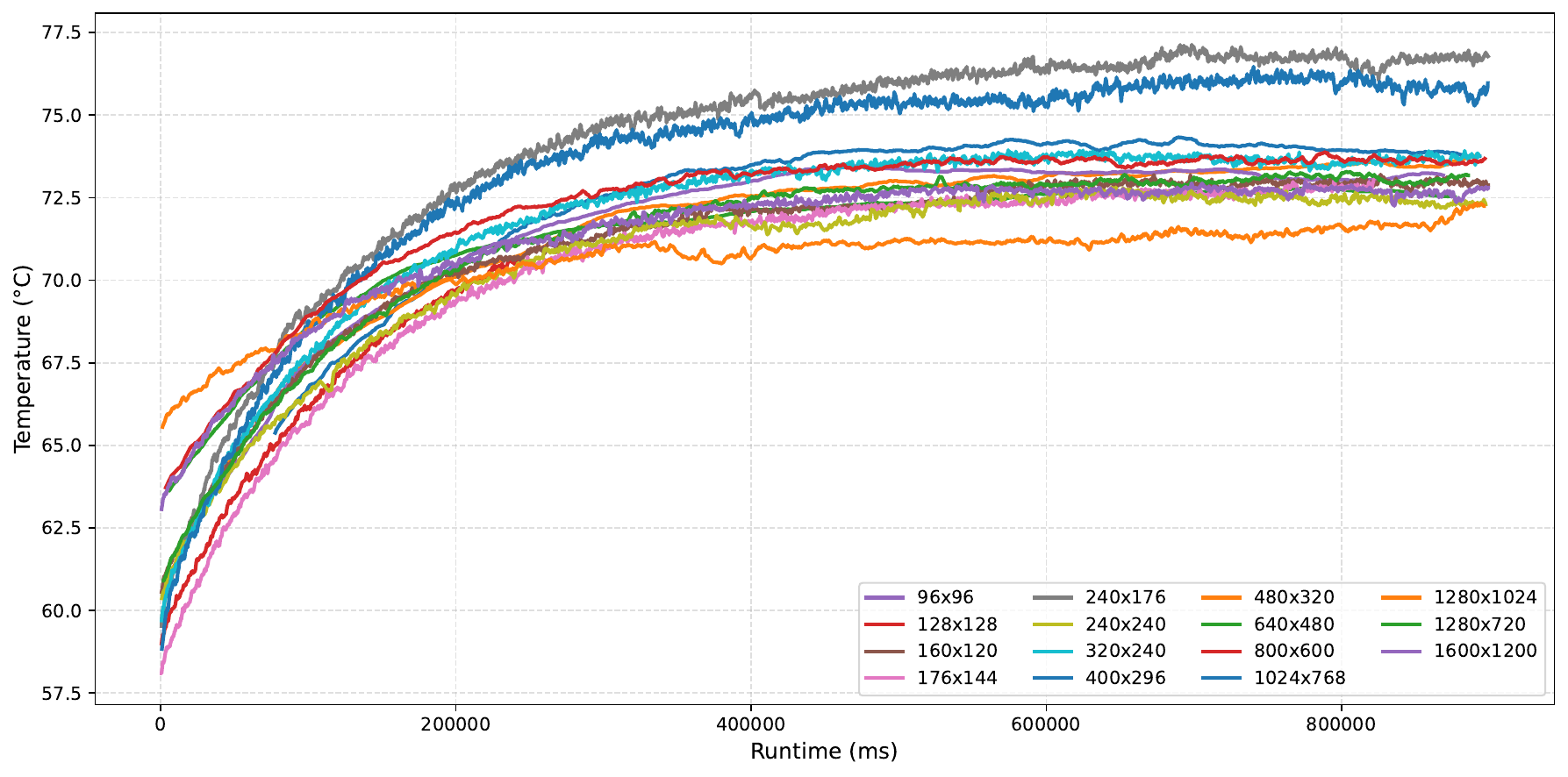}
        \caption{Temperature Deviation across Different Resolutions.}
        \label{fig:temp_deviation}
    \end{minipage}
\end{figure}

Figure-\ref{fig:temp_deviation} indicates that the operating temperature remains relatively stable across all resolutions, ranging from 69.73\degree C to 74.18\degree C. The highest average temperature observed is at 240×176 resolution. The MCU temperature does not exhibit a direct correlation with increasing frame size. 

Figure \ref{fig:frame_size_fps} compares the average FPS and its variability for each resolution. Resolutions below 320×240 consistently achieve FPS above 50. Performance begins to degrade noticeably beyond this point, dropping below 10 FPS at 640×480 and nearing 1 FPS at 1600×1200. The boxplots reveal increasing FPS variability at higher resolutions, highlighting performance instability as computational load increases. This trade-off highlights the need to balance frame size and real-time performance in resource-constrained environments.

Figure \ref{fig:Transmission_vs_res} illustrate the total transmission volume for each resolution over time. System temperature had no observable impact on data throughput. The minimum resolutions suffer from lower speeds and transmission volume due to the smaller size and overhead. Conversely, maximum resolutions are constrained by increased latencies sourced from PSRAM and WiFi. The maximum volume was transmitted for Resolution 400x296, a total of 308.67 MB in 30,493 transmissions over a 15-minute time period. 

\section{Effect of Input Voltage Variation}

\begin{table}[H]
    \centering
    \caption{Performance Metrics for Different Voltages}
    \label{tab:voltage_performance}
    \begin{tabular}{|c|c|c|c|c|}
        \hline 
        \textbf{Voltage} &  \textbf{$\overline{\text{FPS}} \pm \sigma$} & \textbf{$\overline{\text{Temp}} \pm \sigma$ (\degree C)} \\ \hline 
        2v958   & 39.25 $\pm$ 5.51 & 66.39 $\pm$ 1.09 \\ \hline 
        2v98    & 38.87 $\pm$ 5.83 & 66.37 $\pm$ 1.07 \\ \hline 
        3v0     & 38.81 $\pm$ 5.70 & 66.10 $\pm$ 1.81 \\ \hline 
        3v1     & 38.66 $\pm$ 5.88 & 67.85 $\pm$ 1.45 \\ \hline 
        3v2     & 37.03 $\pm$ 6.69 & 67.33 $\pm$ 2.41 \\ \hline 
        3v3     & 40.02 $\pm$ 6.62 & 67.79 $\pm$ 1.71 \\ \hline 
        3v4     & 39.89 $\pm$ 4.84 & 68.21 $\pm$ 2.82 \\ \hline 
        3v5     & 40.19 $\pm$ 5.22 & 67.94 $\pm$ 2.82 \\ \hline 
    \end{tabular}
\end{table}

Table \ref{tab:voltage_performance} presents the thermal behavior of the ESP32-CAM under different input voltages. We observe more deviation and noisy thermal behavior for input voltages over 3 volts. Otherwise, the temperature stabilizes near 65\degree with minimal fluctuation.
 
The highest observed FPS was 40.1 at 3v5, while the lowest was 37.03 at 3v2. Although the FPS wasn't drastically affected by the supply voltage, there is a noticeable trend: voltages above the nominal 3v3 tend to yield slightly higher and stable FPS values, while lower voltages exhibit marginal performance drops. For consistent real-time performance, the power supply should be regulated to avoid dips below this threshold.

\begin{figure}
    \centering
    \begin{minipage}[b]{0.55\textwidth} 
    
        \includegraphics[height=0.45\textwidth]{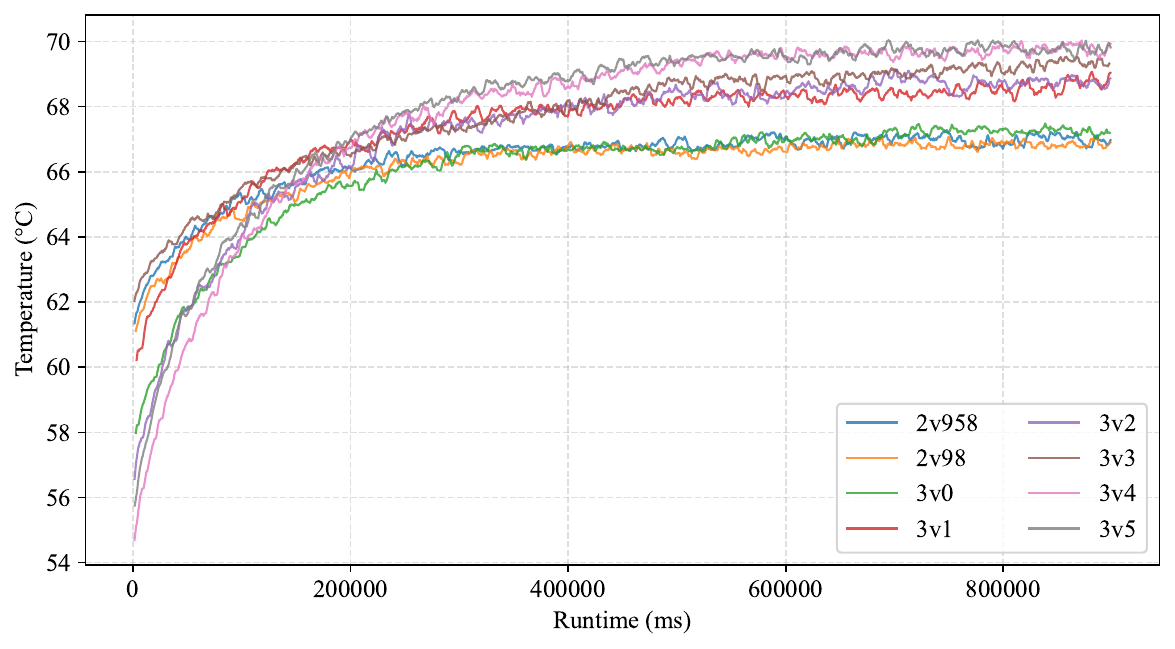}
        \caption{Temperature standard deviation across different voltages.}
        \label{fig:temp_std_dev}
    \end{minipage}
    \hspace{0.02\textwidth} 

    \begin{minipage}[b]{0.24\textwidth} 
        \centering
        \includegraphics[width=\textwidth]{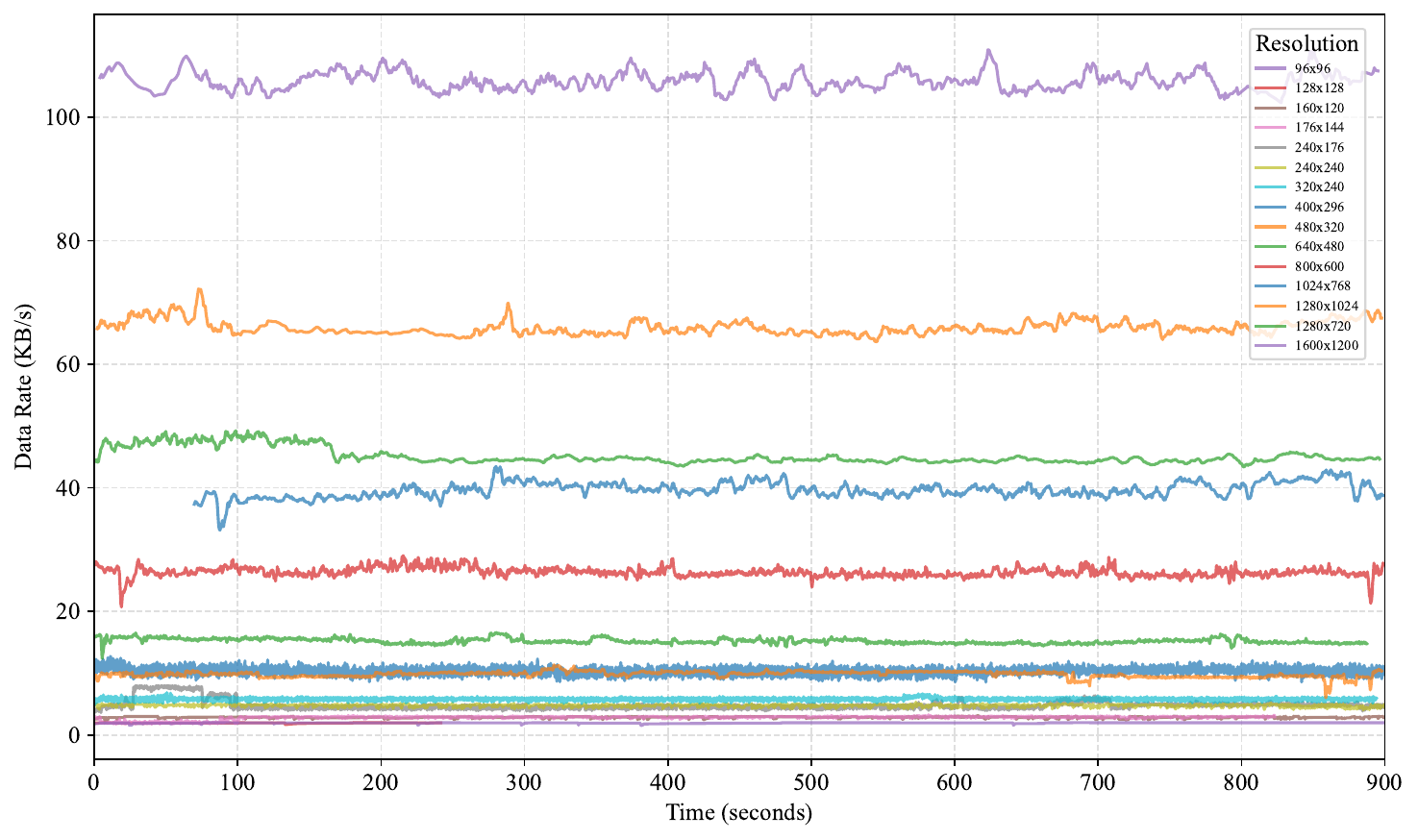}
        \caption{Transmission Rate over time for all resolutions}
        \label{fig:Transmission_vs_res}
    \end{minipage}
    \hfill
    \begin{minipage}[b]{0.24\textwidth} 
        \centering
        \includegraphics[width=\textwidth]{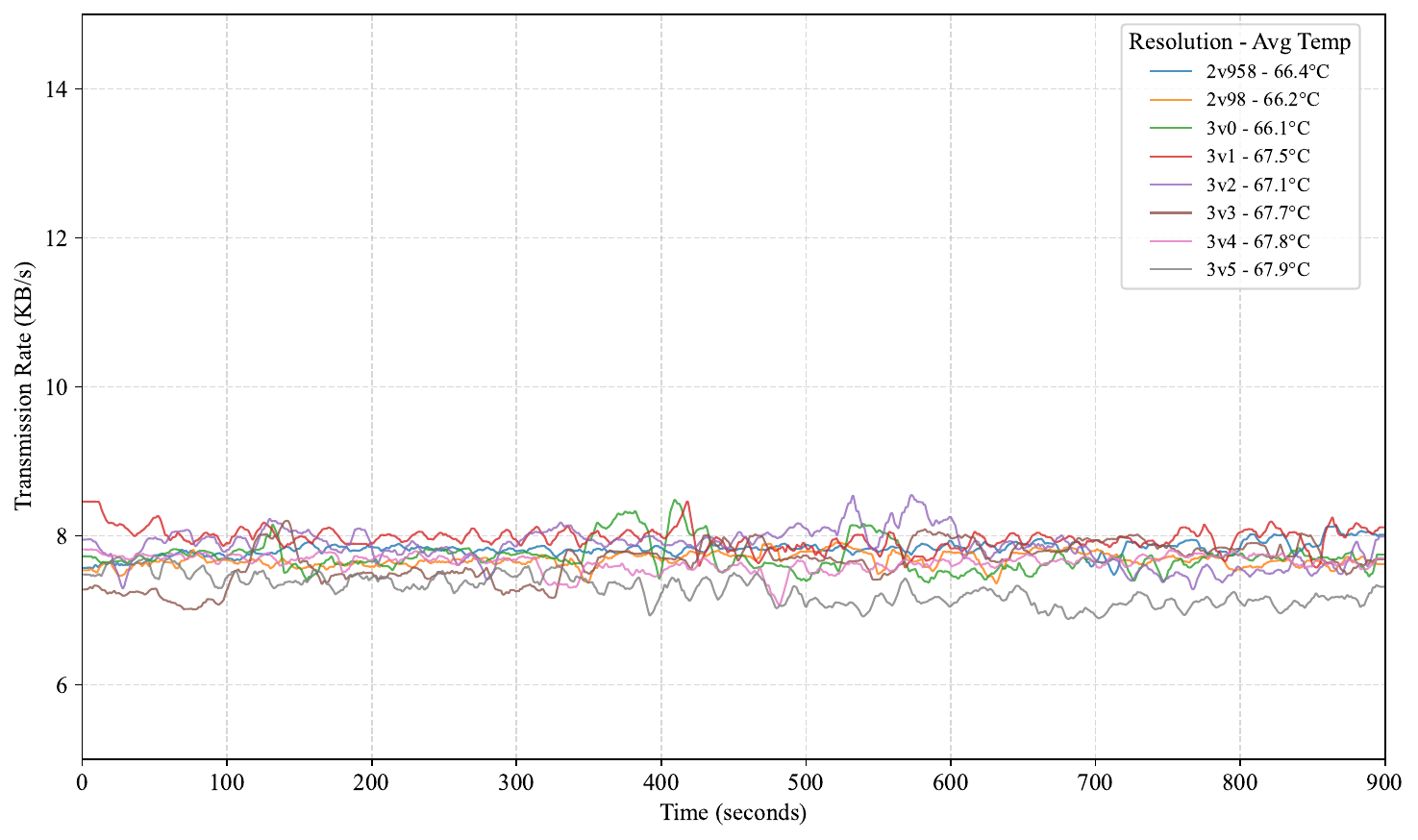}
        \caption{Transmission Rate for varied input voltages(Matched Scale for comparison)}
        \label{fig:transmission_vs_volt}
    \end{minipage}
\end{figure}

Voltage variation had negligible impact on data transmission volume. Figure-\ref{fig:transmission_vs_volt} shows the transmission rates for varying input voltage levels remain stable. The mean bit rate is 7.5 kb/s while the standard deviation is less than 1 kb/s.

\section{Conclusion}
This study provides a detailed performance benchmark for the ESP32-CAM module by addressing and correcting a long-standing bug in the official CameraServer example firmware. Through over six hours of real-time video streaming data across all supported resolutions at different power input levels, we present a comprehensive evaluation of the module’s real-world performance.

Key findings highlight the steep drop in frame rate at higher resolutions due to sensor and memory limitations, while low resolutions consistently deliver over 50 FPS. Thermal behavior remained stable across most operating conditions, and voltage fluctuations had minimal impact on transmission stability unless input levels dropped below 2.985V. This study documents the true performance envelope of the ESP32-CAM and is intended to serve as a reliable reference for engineers, researchers, and developers.

\FloatBarrier
\bibliographystyle{ieeetr}  

\bibliography{references}

\end{document}